\documentclass[sigconf]{acmart}
\usepackage{times}
\usepackage{latexsym}
\usepackage{mathtools}
\usepackage{amsfonts}
\usepackage[ruled,vlined,linesnumbered]{algorithm2e}
\usepackage{microtype}
\usepackage{array}
\title{ReSCo-CC: Unsupervised Identification of Key Disinformation Sentences}

\author{Soumya Suvra Ghosal}
\affiliation{%
  \institution{NIT Durgapur, India}
}
\email{soumyasuvraghosal@gmail.com}

\author{Deepak P}
\affiliation{%
  \institution{Queen's University Belfast, UK}}
\email{deepaksp@acm.org}

\author{Anna Jurek-Loughrey}
\affiliation{%
  \institution{Queen's University Belfast, UK}
}
\email{a.jurek@qub.ac.uk}

\date{}

\copyrightyear{2020}
\acmYear{2020}
\setcopyright{acmcopyright}\acmConference[iiWAS '20]{The 22nd International Conference on Information Integration and Web-based Applications & Services}{November 30-December 2, 2020}{Chiang Mai, Thailand}
\acmBooktitle{The 22nd International Conference on Information Integration and Web-based Applications \& Services (iiWAS '20), November 30-December 2, 2020, Chiang Mai, Thailand}
\acmPrice{15.00}
\acmDOI{10.1145/3428757.3429107}
\acmISBN{978-1-4503-8922-8/20/11}

\begin{document}

\begin{abstract}
Disinformation is often presented in long textual articles, especially when it relates to domains such as health, often seen in relation to COVID-19. These articles are typically observed to have a number of trustworthy sentences among which core disinformation sentences are scattered. In this paper, we propose a novel unsupervised task of identifying sentences containing key disinformation within a document that is known to be untrustworthy. We design a three-phase statistical NLP solution for the task which starts with embedding sentences within a bespoke feature space designed for the task. Sentences represented using those features are then clustered, following which the key sentences are identified through proximity scoring. We also curate a new dataset with sentence level disinformation scorings to aid evaluation for this task; the dataset is being made publicly available to facilitate further research. Based on a comprehensive empirical evaluation against techniques from related tasks such as claim detection and summarization, as well as against simplified variants of our proposed approach, we illustrate that our method is able to identify core disinformation effectively. 
\end{abstract}



\maketitle

\section{Introduction}

The internet is rife with different types of disinformation openly available to the public. Research has shown that disinformation tends to spread much faster and further than truth, as illustrated in a recent study~\cite{vosoughi2018spread}. The spread of fake news is further aided by cognitive patterns such as confirmation bias. Fake news published online can have serious consequences on our health, democracy and economy. Data science approaches for estimating the veracity of an article, i.e., determining whether fake or true, can be broadly seen as exploiting one or more of the following three categories of information: {\it content}, {\it structural} and {\it propagation}. The content refers to the textual as well as any image/multimedia content of the article, whereas structural information refers to the usage of the social network positioning of either the authors or 'sharers' within a social network. Propagation patterns, on the other hand, consider making use of how fast the news is re-shared and whether the sharing happens through specified cross-sections of the network or more broadly. While all three types of information have been used, structural and propagation information have been most popular. In fact, techniques that totally discard content, such as~\cite{ma2017detect} and~\cite{wu2018tracing}, have also met with reasonable success. 

While arriving at a verdict on the veracity of news articles has rightly been the subject of much data science activities in this space, recent guidelines on fact checking, such as that from a recent EU expert group on disinformation~\cite{de2018multi}, place a lot of emphasis on democratic practices on combating fake news such as empowering users through facilitating a positive engagement between users and technologies. A natural direction to tackle this would be to move towards identifying key disinformation extracts from an article as a way of providing more fine-grained information than a single article-level veracity verdict. These extracts, we believe, will function as a prop to encourage the user to ascertain the veracity for herself by carefully perusing such key disinformation. Likely motivated by such considerations, a number of recent techniques adopt {\it claim identification} as an important step~\cite{adler2019real,hassan2017claimbuster,konstantinovskiy2018towards} in the fact-checking process. Our observation has been that disinformation within online articles are not necessarily in the form of well-structured claims, especially in the case of articles comprising long narratives. Further, even when disinformation is embedded within claims, the core disinformation within the article may be localized towards a few claims, than spread evenly across all claims.

%In other words, exposing the rationale behind a veracity decision is key to ensuring that positive societal impact be achieved through fake news detection technologies. 

%Given that content generation outpaces any possibility of manual labelling, automated solutions are required to deal with the disinformation problem. This realization has stimulated a rapid progress in the development of computerised solutions, which allow for a reduction of human effort. Verifying the truthfulness of a content published online is very challenging, particularly so in the case of long articles, which may contain combination of both false and true information intertwined with each other. 

%further indicating the over-reliance on claim-based fact checking may not be very promising. 
%However, long articles contain large number of claims, many of which may not be relevant. With the existing claims detection methods, it is hard to identify which of them are the core claims within the article, which makes the the verification of the article's veracity much more complex.

%One example is the study published in Science Journal, which reported that false information was 70 percent more likely to be retweeted than trustworthy content. 

\subsection{Our Contributions}

In this work, we outline the novel task of {\it identifying sentences containing key disinformation within a long textual article known to contain disinformation}, within an unsupervised setting where no labelled information is available to aid the key disinformation identification process. We propose a technique for the task, called {\it ReSCo-CC}, that adopts a three-step approach. First, sentences in the article are represented within a bespoke feature space, sentence representations being based on their {\it relevance} to the entire article, {\it smoothness} with adjacent sentences, and {\it coherence} as measured using entities referenced within it. Seond, simple existing clustering techniques are applied in order to identify clusters of sentences based on their positioning in the feature space. Third, clusters are ranked on the basis of their {\it coherence} and {\it centrality} in order to identify a cluster of sentences which would be regarded as key disinformation. In addition to a binary sentence-level judgement on whether sentences contain key disinformation as described above, we target to score and rank sentences to reflect their contribution to the overall disinformation within the article. Based on an empirical evaluation against a number of natural baseline methods over a newly curated dataset that is being made publicly available, we establish the effectiveness of our method. 

%We develop a clustering-based core disinformation sentence identification method, that identifies disinformation sentences based on their relation to the other contents in the article. 

%According to our best knowledge, unsupervised identification of core disinformation sentences has not been previously addressed. Our contributions by way of this paper is fourfold. First, we propose the task of key disinformation sentence identification, as a core task towards automating the fact-checking process for long articles. Second, we develop a clustering-based unsupervised methodology that ranks sentences in a document based to reflect their contribution to the overall disinformation within the article. Third, we collect a dataset and propose an evaluation framework that can judge the quality of the key sentence identification algorithm by making use of articles that refute the fake news article. Fourth, we compare against a number of natural baseline methods and empirically illustrate the overall superiority of our methodology. % This is distinct from the task of claims detection in two ways. First, it does not assume that the core sentences must contain a claim. This means that we do not relay on the accuracy of a claim detection algorithm. 

%The paper is organised as follows...

\subsection{RoadMap}

%\noindent{\bf Roadmap:} 
We start with covering some related work in Section~\ref{sec:relwork}. We define our task in Section~\ref{sec:probdef} followed by detailing our proposed method in Section~\ref{sec:method}. Our experimental analysis comparing our method against baselines over a real-world dataset is presented in Section~\ref{sec:experimental}. This is followed by conclusions and pointers for future work. 

\section{Relevant Work}\label{sec:relwork}

As outlined earlier, there has been an abundance of work on automated approaches for fake news detection. In this section, we focus on techniques that attempt to analyze sub-article level information for combating fake news, aligning well with our goal of identifying key disinformation sentences within articles.  The research in the space of sub-article level disinformation spans across multiple fields including ML and NLP, and most of them target one of the following objectives: claim identification, claim verification and verdict justification/explanation. Broadly speaking, the main dimensions on which the existing approaches differ include: (i) claim representation and identification method; (ii) evidence used in the process; (iii) method used for automated fact checking, and (iv) final verdict representation. For (i), claims are often represented as subject-predicate-object triple (e.g. mobile-cause-cancer) or textual fragments. In both cases, the main challenge is the non-trivial level of processing required to extract the claim from a text. The most common approaches rely on a combination of NLP and ML \cite{hassan2017claimbuster}. Within (ii), existing fact checking systems include those that restrict attention to just the information within the claim, and those that seek additional evidence beyond the claim. The former estimate the veracity of a claim using surface-level linguistic features \cite{rashkin2017truth} and additional metadata like author’s history of previous false claims \cite{wang2017liar}. The fact checking task itself (i.e., (iii)) is often defined as a supervised learning problem. It involves construction of a text classification model using labelled data, such as existing claims previously annotated as fake or non-fake. The classifier is further applied to assign fake/non-fake label to a new claim \cite{wang2017liar,rashkin2017truth} or to detect fact-check worthy claims \cite{hassan2017toward}. Text classification only considers features extracted from the claim itself or the author’s profile and does not rely on any other sources of evidence. Evidence based methods commonly consider fact checking as a knowledge graph analysis problem in order to predict whether an unobserved triple is likely to appear in the graph \cite{ciampaglia2015computational,vlachos2014fact,thorne2017extensible}. Another common approach casts fact checking as a textual entailment task, modelled to predict whether a document or its part is for, against or observing a given claim \cite{ferreira2016emergent,riedel2017simple}, or in order to retrieve sentence-level evidence for the claim ~\cite{hua2017understanding}. Methods that rely on repositories of previously fact checked claims are mainly based on sentence-level textual similarity \cite{vlachos2014fact,hassan2017claimbuster}. Task (iv), in the simplest scenario, is just about providing a true/false output directly from a binary classification engine \cite{nakashole2014language}. Alternatively, multi-class labels over a multi-point scale~\cite{rashkin2017truth}, or a numerical range indicating how likely a claim is to be true \cite{bast2017overview} are used. 

Against the backdrop of claim-based and NLP-oriented data science solutions, we propose a significantly different task formulation, that of scoring sentences based on their contribution to the overall article disinformation. While we haven't come across work addressing this particular task, we will evaluate our method against state-of-the-art claim detection and summarization methods. Our task is designed to be invoked on only those news articles that are known to contain disinformation; thus, identifying disinformation-laden articles is a task that is upstream to ours. In addition to a variety of supervised learning methodologies that seek to accurately identify articles as containing fake news or not, there has been recent interest in unsupervised fake news detection as well. Such unsupervised methods widely varying in character, with techniques capitalizing on synchronous user behavior~\cite{DBLP:conf/ht/Gangireddy0L020} and emotions~\cite{DBLP:conf/ideas/K0L20}. Other more subtle features such as lexical~\cite{inra2020lex} and thematic~\cite{inra2020them} have also been found to be useful to differentiate fake news from real news without the aid of supervision.

\section{Task Definition}\label{sec:probdef}
\label{sec:taskdef}
%Consider a collection of documents $D={d_1,...,d_n}$ which content represents some disinformation. Each document $d_i$ contains $m_i$ sentences. Accordingly, we can represent a document $d_i$ as $[s_{i1},...,s_{im_{i}}]$. Furthermore, we aim for a numerical score to be assigned to each sentence, which is directly related to how much this sentence contributes to the disinformation of the document. 

Consider a document $D$ which is known to contain some disinformation. Let the document comprise $n$ sentences; $D = [ s_1, s_2, \ldots, s_n ]$. Our task targets unsupervised scoring of sentences {\it at a per-document level}, in a way that the scoring correlates with the contribution of the sentence to the overall disinformation within the document. %Notationally:
%\vspace{-0.5em}
\begin{equation}
%\vspace{-0.5em}
\big[ D = [{s_{1}, \ldots ,s_{n}}] \big] \xRightarrow[\text{learning}]{\text{unsupervised}} [ r_1, \ldots, r_n ]
\label{eq:task}
\end{equation}

We would like to consider two scenarios, with each $r_i$ to be either a binary or numeric value indicating whether/how much $s_i$ contributes to the disinformation within document $D$. We call the setting of binary $r_i$ as the {\it identification} task, and the numeric $r_i$ setting as the {\it scoring} task. Unlike the formulation of similar tasks such as claims detection which are typically formulated within supervised settings, we address the unsupervised identification/scoring problem. %Second, we do not make assumption that only sentences that are claims can contain disinformation. Third, we are able to identify the top sentences representing the disinformation within a document. 

\subsection{Motivation and Positioning the Task}

As outlined earlier, we start with the assumption that the article in question, $D$, contains disinformation. This task is thus designed to be downstream to the disinformation identification task that determines whether or not a news article contains disinformation. Once an article has been identified to contain disinformation, the simplest strategy to present it to the user would be using a warning pop-up, or in cases where the disinformation identification is implemented in a browser plugin, to issue a warning before displaying the article webpage. This is analogous to the usage of warnings in Twitter\footnote{https://www.bbc.co.uk/news/technology-52846679}, where a tweet of suspected credibility is hid behind a warning. As may be obvious, articles identified not to contain disinformation will not be intercepted in any way. However, as noted in the EU report~\cite{de2018multi}, such a paradigm of presenting disinformation risks undermining freedom of expression as well as the diversity of the news media ecosystem, and democratic liberal values in general. The task of key disinformation identification offers a pathway towards addressing this issue, that of enabling selective highlighting of key disinformation sentences in an article. Such selective highlighting, with appropriate indicators, as well as perhaps shortcut links to authoritative information on the topic of the sentence, would align the presentation paradigm with a much more democratic model focusing on user empowerment and positive engagement (as alluded to earlier). Thus, the key disinformation identification task enriches the fake news detection pipeline with presentation opportunities in line with cherished democratic values. Any improvements achieved in the upstream task of disinformation identification will evidently be compelementary to this task, and would enrich the upstream part of the pipeline. 

\section{ReSCo-CC: Our Method}\label{sec:method}

%The goal of our method is twofold. For a given disinformation-laden article, we want to {\it identify} which sentences are associated with the overall disinformation that the document contains. {\it Second}, we want to be able to {\it score those sentences} based on how much they contribute to the disinformation within the document. 

Our method, codenamed {\it ReSCo-CC}, targets to address both the identification and scoring tasks outlined in Section~\ref{sec:taskdef}. Our approach consists of three distinct phases; (i) representing sentences in a bespoke feature space, (ii) clustering the sentence representations, and (iii) choosing sentences based on clustering output. We describe these in separate subsections herein. 

\subsection{Bespoke Feature Space for Sentences}

We undertook a qualitative study of various fake news articles as well as literature on fake news from journalism with a view of understanding the aspects that could contribute to sentences within it being regarded as core disinformation. Our focus was on identifying sentence characteristics indicative of core disinformation content that may be considered as sufficiently generic - so it would not need additional fine-tuning to work for niche sub-domains - with an additional preference towards characteristics that are amenable to computational modelling. Our analyses indicated that there are at least three kinds of sentences within a disinformation-laden document; sentences that contain key disinformation, sentences that cushion the key disinformation through providing (tangential) truthful information, and sentences that are oriented towards smoothening the overall flow. Based on this observation, we distilled three key sentence characteristics from our study, viz., {\it relevance}, {\it smoothness} and {\it coherence}, abbreviated together as {\it ReSCo}, which forms the basis of our bespoke feature space for key disinformation identification. These three features bear significant complementarities, {\it relevance being a document-level feature}, {\it smoothness being a more fine-grained feature estimated at a sub-article level}, whereas {\it coherence is a sentence-level feature assessed in reference to a knowledge base}. We describe them herein. 

\subsubsection{Relevance.}

Consider each sentence $s_i$ as being represented using a vector $v_i$. While any sentence embedding could be used for $v_i$, we model $v_i$ as the average of the pre-trained Google word2vec word-level embeddings\footnote{https://code.google.com/archive/p/word2vec/}. Now, the relevence of each sentence wrt $D$, denoted as $rel(s_i)$ is calculated as:

\begin{equation}
rel(s_i) = \frac{1}{n-1} \times \sum_{j \in \{1 \ldots n\}, j \neq i} cosine(v_i, v_j)
\label{eq:rel}
\end{equation}

Thus, the relevance is quantified as the average of sentence level similarities between $s_i$ and each other sentence in $D$, estimated using the chosen vector representation. Under this formulation, which has a flavor of document-level summarization, sentences that bear {\it higher} similarities with a {\it larger} number of $D$'s sentences will be scored higher. In other words, sentences that are semantically central to the document are likely to achieve higher $rel(s_i)$ scores. 

\subsubsection{Smoothness.}

Smoothness measures how well each sentence gels with sentences on either side of itself. We quantify it using the same vector representation $v_i$ (as introduced above) as:

\begin{equation}
smo(s_i) = \begin{cases}
cosine(v_i, v_{i+1}) & i = 1 \\
\frac{1}{2} \times \big( cosine(v_{i-1},v_i) + cosine(v_i, v_{i+1}) \big) & 1 < i < n \\
cosine(v_{i-1},v_i) & i=n
\end{cases}
\label{eq:smo}
\end{equation}

Thus, this is a local feature, quantified using the similarity between each sentence $s_i$ and it's adjacent ones. Intuitively, it measures how well $s_i$ maintains the local {\it 'flow'} of information in the document, with those that offer a smooth flow being accorded higher $smo(.)$ scores. 

\subsubsection{Coherence.}

Unlike the above two measures, coherence measures the coherence of a sentence in reference to an external knowledge base. First, we identify mentions of Wikipedia entities within each sentence using entity linking methods~\cite{gupta2017entity}. We denote the set of entities identified within a sentence as $E(s_i)$. We then use the Wikipedia2Vec method~\cite{yamada2018wikipedia2vec} in order to map each entity $e \in E(s_i)$ to a vector, denoted as $v_e$. We regard coherent sentences as those that refer to a number of related entities, yielding the following formulation for coherence:

\begin{equation}
coh(s_i) = \frac{\sum_{e, e' \in E(s_i), e \neq e'} cosine(v_e, v_{e'})}{\sum_{e, e' \in E(s_i), e \neq e'} 1}
\label{eq:coh}
\end{equation}

Thus, $coh(s_i)$ measures the average of pairwise similarities between entities mentioned within $s_i$. As a feature of $s_i$ independent of all other sentences in $D$, this stands very complementary to the relevance and smoothness features. 

\begin{table*}[t]
\caption{Illustration of features on a sample fake news article represented as an array of sentences in the lexical order within the document. The sentences that score the highest on Relevance ({\bf Rel}), Coherence ({\bf Coh}) and Smoothness ({\bf Smo}) are illustrated in the table.}
\label{tab:features}
\centering
\begin{tabular}%{|c|c|}
	{|c|p{10cm}|}
	\hline
	 & {\small Eczema is the most common skin disease worldwide.} \\
	\hline
	& {\small A new clinical trial is testing a natural treatment that researchers hope will provide a long-term solution for those dealing with the dry, itchy and painful skin that comes with chronic eczema.} \\
	\hline
	& {\small The trial uses a cream containing beneficial bacteria to fight harmful bacteria on the skin. }\\
	\hline
	 {\bf Smo} & {\small While it may seem counterintuitive to treat bacteria with more bacteria, experts say this approach seeks to restore the natural microbial balance of healthy skin.} \\
	\hline
	& {\small "There are over 1,000 species of bacteria that all live in balance on healthy skin, some that even produce natural antibiotics." }\\
	\hline
	& $\ldots$ \\
	\hline
	{\bf Coh} & {\small Powerful antibiotics are commonly prescribed for eczema, but they kill good bacteria on patients' skin along with the bad. }\\
	\hline
 	& $\ldots$ \\
	\hline
	{\bf Rel} & {\small Experts say there is more research to be done, but that the goal of the trial is to discover the best combination of bacteria to clear eczema from the skin and then make it available to patients as a prescription cream.} \\
	\hline
\end{tabular}
%\vspace{-0.2in}
\end{table*}

\subsubsection{Illustration.}

We now illustrate the qualitative difference between the above three features using extracts from a fake news article in the dataset we use in our experiments, in Table~\ref{tab:features}. As maybe noted therein, the top-scoring sentence on each for the features are quite different qualitatively. The top-Rel sentence is seen to be quite central to the article, whereas the top-Smo sentence identifies a region of very smooth flow of text, whereas the top-Coh sentence is seen to talk about very related entities (e.g., antibiotics and bacteria). We do not imply that the top-scored sentences are likely to be key disinformation, but simply that these are useful and complementary indicative features to capture; more details on using {\it ReSCo} for key disinformation identification and the motivation on using the three features will follow. %As we will outline in a later section, a set of sentences that cohesively offer a balance along these features is the eventual heuristic that our method rests on. 

\subsection{Clustering}

Having represented the sentences within $D$ in the $\mathbb{R}^3$ {\it ReSCo} space as above, we now cluster $D$ to form groups of sentences that are coherent in the $\mathbb{R}^3$ {\it ReSCo} space. This may be accomplished by any clustering algorithm; we use the popular centroid-based partitional clustering algorithm, $K$-Means~\cite{macqueen1967some}. $K$-Means requires a parameter, the number of clusters in the output. Given that individual articles may vary much in the number of sentences they contain, we use the elbow method~\cite{kodinariya2013review} to choose the value of $K$. The elbow method is a popular method of choosing the number of clusters in the data in a mathematically principled manner~\cite{tibshirani2001estimating} that involves choosing a trade-off between competing criteria of fewer clusters and more coherent ones; in particular, it chooses the point beyond which returns of choosing larger numbers of output clusters saturate.

%, and also employ the $K$-Means++ initialization heuristic~\cite{arthur2006k}

%\footnote{https://www.geeksforgeeks.org/elbow-method-for-optimal-value-of-k-in-kmeans/}

\subsection{Identification and Scoring of Sentences}

It would rightly appear that each of the {\it ReSCo} features are correlated with what may be regarded as {\it readability} and {\it quality}, and one may be left wondering how such a set of features would lend to disinformation identification. This brings us to the key intuition and heusitic within our method. While relevance, smoothness and coherence are all individually {\it good characteristics}, the combination poses trade-offs due to some conflicting elements across them. A good quality article, i.e., one that is crafted to deceive the reader into believing the disinformation, would comprise sentences, each of which would excel in any one, or perhaps two, of those features. For example, a sentence that delves into some minor important points in the overall narrative would suffer on {\it relevance}, and sentences that connect different sub-narratives would suffer on {\it smoothness} and {\it coherence}. In other words, good quality articles encompass a bouqet of sub-narratives held closely together, making individual sentences unable to optimize on all {\it ReSCo} features. On the other hand, two key observations from our qualitative analysis guide the formulation of our identification/scoring method. First, we observe that fake news authors ensure that key disinformation sentences are not particularly disadvantaged on any of the {\it ReSCo} features; this is implicitly due to the urge to optimize on {\it readability} and {\it quality}. Given that optimizing for all are quite challenging, this leads to key disinformation sentences being positioned close to the centroid of $D$'s {\it ReSCo} space. Secondly, the positioning in the {\it ReSCo} space is strongly influenced by the style of the author and the topic in question, making it likely that the key disinformation sentences are all very close to each other in the {\it ReSCo} space. These lead to our two key heuristics, {\it Centrality} and {\it Cohesion}, for identification and scoring of key disinformation sentences. With cohesion being enforced by the clustering step, the key disinformation sentences are those belonging to the cluster whose centroid is closest to $D$'s {\it ReSCo} space centroid; this cluster is denoted as $C^*$:

%our observation has been that fake news authors position key disinformation sentences such that they moderately optimize for each of the {\it ReSCo} features, and do so in a very similar fasion. 

%On the other hand, fake news authors cannot risk sacrificing quality and readability,  being largely determined by those factors. The mutual conflicts between {\it ReSCo} features would limit optimizing all of them together, leading to a positioning where each of them are optimized for in a balanced way; this translates to an expectation that such key disinformation sentences may be close to the centroid of $D$'s {\it ReSCo} space. Further, based on the positioning of the author's writing style in the {\it ReSCo} space, we expect that key disinformation sentences will all be close enough to each other clustered around that position. 

\begin{equation}
C^* = \mathop{\arg\min}_{C \in \mathbb{C}}\ \ \ eucl(avg\{ReSCo(s)|s \in C\}, avg\{ReSCo(s)|s \in D\})
\label{eq:cstar}
\end{equation}

where $\mathbb{C}$ is the set of all clusters output from the clustering step, $ReSCo(s)$ being the 3-d {\it ReSCo} space representation of sentence $s$, and $eucl(.)$ denoting the euclidean distance. The binary output for the identification task is computed as:

\begin{equation}
r_i = \begin{cases}
1 & s_i \in C^* \\
0 & otherwise \\
\end{cases}
\label{eq:ribin}
\end{equation}

In other words, the {\it identification} task is accomplished by choosing all sentences in the cluster whose {\it ReSCo}-space centroid is closest to that of all sentences within $D$. The $r_i$ output for the {\it scoring} task computes a score for all sentences in $C^*$ based on its proximity to $C^*$'s centroid. Sentences belonging to the other clusters are left at $0.0$; this is shown as below:

\begin{equation}
r_i = \begin{cases}
cosine(ReSCo(s_i),avg\{ReSCo(s)|s \in C^*\}) & s_i \in C^* \\
0.0 & otherwise \\
\end{cases}
\label{eq:riscore}
\end{equation}

The scoring provides a way to discriminate among sentences in $C^*$; especially, if $C^*$ contains many sentences and a top-$k$ needs to be chosen, the selection may be done using the scoring in Eq.~\ref{eq:riscore}. This completes the description of our method, codenamed {\it ReSCo-CC}, that uses {\it ReSCo} features and the {\it C}entrality and {\it C}ohesion heuristics to perform key disinformation identification.

\subsection{ReSCo-CC: The Overall Method}

The overall method is outlined in Algorithm~\ref{alg:algo}. The first two steps correspond to the design of the {\it ReSCo} space, with the clustering in Step 3 followed by the identification or scoring process in Steps 4-5. Once the {\it ReSCo} embedding is achieved, all further steps are linear. The computation of $rel(.)$ in achieving the {\it ReSCo} embedding, however, is quadratic in the number of sentences in the document. Given that typical documents contain only 15-30 sentences, the quadratic complexity does not pose much concern. {\it ReSCo-CC} was found to offer turn around times of milliseconds even in quite long documents. 

\begin{algorithm}[t]
\SetKwInOut{Input}{input}
\SetKwInOut{Output}{output}
%\SetKwInOut{Initialization}{Initialize}
\Input{Document $D=[s_1,...,s_n]$}
\Output{$\{ \ldots r_i \ldots\}$ for the scoring or identification task}
$\forall\ i, v_i \leftarrow GetWord2Vec(s_i)$ \\
$\forall\ i, ReSCo(s_i) \leftarrow < rel(s_i), smo(s_i), coh(s_i) >$ (Eqs~\ref{eq:rel},\ref{eq:smo},\ref{eq:coh}) \\
$\mathbb{C} \leftarrow KMeans(\{ \ldots, ReSCo(s_i), \ldots \})$ \\
Determine $C^*$ according to Eq.~\ref{eq:cstar}\\
Determine $r_i$s according to Eq.~\ref{eq:ribin} or Eq.~\ref{eq:riscore} and output.
%Output $r_i$s 
%%\For{$s_i \in D$}{%
%%     $v_i \leftarrow GetWord2Vec(s_i)$
%%    }
%$S$=$\emptyset$
%\\\For{$s_i \in D$}{%
%     $<f^1_{s_i},f^2_{s_i},f^3_{s_i}> \leftarrow$ $rel(s_i), smo(s_i), coh(s_i)$
%%     $f^1_{s_i} \leftarrow relevance$\\%\dfrac{1}{n-1}\sum_{s_j \in D\setminus s_i} Cos(s_i,s_j)$
%%     $f^2_{s_i} \leftarrow smoothness$\\%\dfrac{1}{2}(Cos(embd(s_{i-1}),embd(s_i)),Cos(embd(s_{i+1}),embd(s_i)))$
%%     $f^3_{s_i} \leftarrow coherence$
%     \\$S=S \cup <f^1_{s_i}, f^2_{s_i}, f^3_{s_i}>$
%    }
%${(C_i,c_i),...,(C_k, c_k)} \leftarrow kMeans(S)$
%\\$\overline{s} = \dfrac{1}{n}\sum_{i=1...n}s_i$
%\\$\overline{C}=\underset{C_i}{\mathrm{argmin}}$ $Euclidean(c_i,\overline{s})$
%\\$\overrightarrow{V_n}=<0,...,0>$
%\For{$i=1...n$}{%
%	\If{$s_i \in \overline{C}$}{$\overrightarrow{V_n}(i)=1$}
%	}
%$R \leftarrow Rank(\overline{C})\, by\, Euclidean(s_i,\overline{c})$	
%\\Output $\overrightarrow{V_n}, R$ 
\caption{ReSCo-CC}
\label{alg:algo}
\end{algorithm}
\section{Experimental Evaluation}\label{sec:experimental}

We describe the experimental setup followed by results and analysis. 

\subsection{Experimental Setup}

\subsubsection{Creation of gold standard dataset}

Evaluation of the key disinformation sentence identification task would ideally require sentence level manual labellings. Being a novel task, we have not come across such sentence annotated datasets. Further, we found that health domain fake news often requires signifciant domain expertise to annotate, something that we did not have access to. Thus, we develop an innovative approach for generating labelled data for evaluating our approach. We identified a website, {\it healthnewsreview.org}, which targets, among other initiatives, to debunk fake health news ({\it hoaxes}, as they call it). They produce, for each fake health news piece, a textual refutation of the same detailing why the information in the newspiece is false. {\it We selected $80$ articles which were debunked within {\it healthnewsreview}, to form our empirical testbed. It may be noticed that ours is an article-level task, thus each article is treated separately, making the small size of the dataset not an overwhelming concern, not as much for scenarios that do corpus level learning.} We use simple statistical methods over the dataset comprising $(hoax, refutation)$ pairs, to arrive at a disinformation scoring for each sentence in the hoax article. For each sentence $s$ in the hoax (the hoax file is also the same that will be input to the method as $D$), the score is as follows:

% in the space of critical thinking about healthcare

\begin{equation*}
score(s) = \\ avg \{ cosine(word2vec(s),word2vec(r)) | r \in refutation \}
\end{equation*}

This estimates $score(s)$ as the average similarity that sentences in the refutation have, with $s$. The refutation narratives in {\it healthnewsreview} were found to focus on the key disinformation sentences in the hoax. Given that, we expect that $score(s)$ estimated as average similarities to the refutation sentences to approximate the disinformation-ness of each sentence; we verified this assumption through manual perusal of a significant fraction of the dataset. Thus, high $score(.)$ would be associated with sentences that hold key disinformation. We call this $\{ \ldots, score(s_i), \ldots \}$ vector as the {\it refsim} vector, short for {\it refutation similarity}. This vector holding similarities to the refutation forms our gold standard labelling for experiments; as may be obvious, the refutations and thus the {\it refsim} vector are not available to the methods, and are used only in evaluation. {\it We are making this dataset - comprising both the (hoax, refutation) pairs and the refsim vectors - available in the public domain, at \url{http://member.acm.org/~deepaksp} > Publications > Entry for ReSCo-CC paper.}

\subsubsection{Evaluation metrics.}

We measure the effectiveness of the binary and scored versions of the $r_i$ estimates from {\it ReSCo-CC} against the $refsim$ vector using two evaluation models. Firstly, for the identification setting, our task is to compare the $r_i$s which are in $\{0,1\}$ with the numeric $refsim$ vector. As a natural method for comparing two vectors, we use the Pearson product moment correlation co-efficient. Pearson correlation co-efficient has been popular in evaluating NLP methods, and has been extensively used in evaluating methods in shared tasks such as SemEval~\cite{cer2017semeval} and others\footnote{https://saifmohammad.com/WebPages/EmotionIntensity-SharedTask.html}. Second, for the scoring task, our intent is on understanding whether the high $r_i$ values indeed correlate with the highest scored sentences from $refsim$. Towards bringing this into a standard information retrieval evaluation model, we discretize/truncate the gold-standard $refsim$ vector by retaining the top-$\rho$ sentences according to $score(.)$ as {\it key disinformation} and others as not, forming a binary labelling. Now, the ranking offered by $r_i$s is evaluated using NDCG~\cite{wang2013theoretical} to measure whether they rank the key disinformation (i.e., top-$\rho$) sentences highly enough. NDCG is the de facto measure for evaluating ranking quality in information retrieval (an analysis of its theoretical underpinnings appears at~\cite{wang2013theoretical}); the construction of NDCG accounts for not just whether the expected results appear in the top-$k$, but also quantifies how close to the top the correct results appear. We also vary the $refsim$ discretization/truncation parameter $\rho$ to study the effectiveness trends. For both these metrics, higher values are desirable and correlate with better effectiveness. Each of these are computed at the document level; we report the average of these measures over the $80$ documents in our dataset.

%We used two metrics to evaluate the proposed method. First one is the Pearson product moment correlation co-efficient between the binary vectors returned by {\it ReSCo-CC} (or a baseline method) and the {\it refsim} vector. Second one is the NDCG measure~\cite{jarvelin2002cumulated} used to compare the ranking of sentences returned by our method against a gold-standard labelling set to one of top 3, 5 and 7 sentences according to {\it refsim} scoring. 

\subsubsection{Baselines and Setup.}

Given that our task is novel, there exist no methods in literature addressing the precise task. Thus, we adapt methods for related tasks to serve as baselines in our empirical evaluation. Our task of key disinformation sentence identification/scoring has a similar structure in construction to {\it claim detection} (typically supervised) and {\it document summarization} (typically unsupervised, like our task). Algorithms for both tasks are capable of operating in the binary output model (as in our identification task) and the scored model (as in our scoring task). Thus, we use state-of-the-art techniques from these tasks as baseline methods; these are (i) a fact-checking oriented supervised claim detection method~\cite{adler2019real} (we abbreviate it to {\it CD}) trained over IBM Debator dataset, and (ii) the deep learning-based document summarization method from a very recent paper~\cite{dong2019unified} that has been shown to produce good document summaries. In addition to these, we also evaluate against variants of our method formed by dropping each of the {\it centrality} and {\it cohesion} assumptions, to arrive at an ablation study. First, we omit centrality and focus on cohesion by choosing the most coherent cluster, the one that has the highest average pairwise similarity between its members; we call this {\it ReSCo-Coh}. Similarly, we relax the cohesion criterion and rank sentences based on their similarity to the dataset centroid along the {\it ReSCo} features; this is called {\it ReSCo-Cen}. In order to achieve robust results (given the randomness in the clustering initialization step), we report results averaged over $100$ iterations. 

%uses the Google Universal Sentence Encoder

\subsection{Results and Analysis}

Tables \ref{tab:Results1} and \ref{tab:Results2} list the results of the comparative evaluation of our method against the four baseline techniques based on the evaluation measures of {\it correlation co-efficient} and {\it NDCG} respectively. It may be noted from Table~\ref{tab:Results1} that {\it ReSCo-CC} achieves significantly higher performance than the baselines, with the second best one, {\it CD} (which is notably a supervised technique, being trained over the IBM Debator dataset), being left far behind. These trends continue over into Table~\ref{tab:Results2} as well, across varying values of $\rho$. Across both these settings, we carried out the two-tailed t-test\footnote{https://en.wikipedia.org/wiki/Student\%27s\_t-test}, wherein it was observed that {\it ReSCo-CC gains over each of the baselines were staistically significant} with a p-value < 0.01. While all techniques show an increase with increasing $\rho$, it may be noted that NDCG is more meaningful for low values of $\rho$. For example, at the extreme case, setting $\rho = n$ is equivalent to considering all sentences in the document as key disinformation sentences, making it meaningless since all techniques would score $NDCG=1.0$ by design. Given this construction of the evaluation, we find it very promising to note that {\it ReSCo-CC} scores 0.856 even at a relatively low value of $\rho=3$. In addition to the improvements over {\it CD} and {\it Summ}, it is seen that {\it ReSCo-CC} outperforms both {\it ReSCo-Coh} and {\it ReSCo-Cen} by very large margins, indicating that the combination of clustering and cohesion is very pertinent to core disinformation detection within our method, and relaxing either one would lead to significant drops in accuracy. 

%It can be noted that according to each of the evaluation measures, our method obtained the best results. We can observe that the modified version of our method performed significantly worse. The most competitive turned out out be the claim selection technique, however, it is still significantly behind our method.
\begin{table}[t]
\caption{Evaluation based on Correlation Co-efficient.}
\vspace{-1em}
\label{tab:Results1}
\centering
\begin{tabular}%{|c|c|c|}
	{p{2cm}p{2cm}p{1.5cm}}
	\hline
	\hline
	{\bf Method} & {\bf Mean} & {\bf Std Dev}\\\hline
	\hline
	{ReSCo-CC} & {\bf 0.851} & {0.115}\\ %\hline
        {CD} & 0.612 & 1.29 \\%\hline
        {Summ} & 0.218 & 0.084 \\%\hline
        {ReSCo-Coh} & 0.400 & 0.067 \\%\hline
        {ReSCo-Cen} & 0.394 & 0.15 \\\hline \hline
\end{tabular}
%\vspace{-1em}
\end{table}
\begin{table}[t]
\caption{Evaluation based on NDCG using the top-$\rho$ sentences from {\it refsim} as relevant, with $\rho \in \{3,5,7\}$.}
\vspace{-1em}
\label{tab:Results2}
\centering
\begin{tabular}%{|c|c|c|c|}
	{p{2cm}p{1.3cm}p{1.3cm}p{1.3cm}}
	\hline
        \hline
	{\bf Method} & {\bf $\rho$=3} & {\bf $\rho$=5} & {\bf $\rho$=7}\\\hline \hline
	{ReSCo-CC} & {\bf 0.856 } & {\bf 0.880} &{\bf 0.910}\\ %\hline
        {CD} & 0.487 & 0.690 & 0.721 \\%\hline
        {Summ} & 0.246 & 0.294 & 0.315 \\%\hline
        {ReSCo-Coh} & 0.320 & 0.380 & 0.450 \\%\hline
        {ReSCo-Cen} & 0.368 & 0.420 & 0.498 \\ \hline \hline
\end{tabular}
\vspace{-1em}
\end{table}

\begin{table*}[t]
\caption{Illustration of {\it ReSCo-CC} result for a sample document, with excerpts shown, along with {\it refsim} scores. }
\label{tab:results}
\centering
\begin{tabular}%{|c|c|}
	{|c|p{9cm}|c|}
\hline
{\it ReSCo-CC} & \multicolumn{1}{c|}{{\it Sentences from the Article}} & $refsim$ \\
{\it Output} & & \\
\hline
 0 & {\small About 3 million people in the US are diagnosed every year with bipolar disorder, a psychiatric condition characterized by dramatic shifts inood from depression to mania.} & 0.50 \\
\hline
0 & {\small Currently, the standard treatment includes a combination of psychotherapy and prescription medications such as mood stabilizers and antipsychotics.} & 0.52 \\
\hline
1 & {\small However, an emerging field of research is exploring the use of probiotics--often thought of as "good bacteria"--as a potential new avenue for treatment of bipolar and other psychiatric mood disorders.} & 0.63 \\
\hline
1 & {\small And a new study from Baltimore's Sheppard Pratt Health System, conducted by a research team led by Faith Dickerson, finds that a probiotic supplement may reduce inflammation of the gut, which is known to exacerbate bipolar disorder.} & 0.66 \\
\hline
0 & {\small Probiotic organisms are non-pathogenic bacteria that, when present in the gut flora, are known to improve the overall health of the host.} & 0.56 \\
\hline
0 & {\small Studies have shown that the intimate association between the gut microbiome and GI tissue has a significant effect on the GBA.} & 0.56 \\
\hline
1 & {\small There is also mounting evidence linking imbalances in the microbial species that make up the gut microbiome to a number of health problems including allergies, autoimmune disorders, and psychiatric mood disorders.} & 0.60 \\
\hline
1 & {\small In the case of bipolar disorder and the GBA, previous studies have shown that inflammation, or overstimulation of the body's immune system, is a contributing factor in the disease.} & 0.62 \\
\hline
$\ldots$ & \multicolumn{1}{c|}{$\ldots$} & $\ldots$ \\
\hline
\end{tabular}
\end{table*}

\subsection{Qualitative Analysis}

Table~\ref{tab:results} illustrates the {\it ReSCo-CC} outputs on a part of a sample document; we have chosen eight sentences from a part of the document with key disinformation for illustration. The first column indicates the binary identification output from {\it ReSCo-CC} for the sentence in the second column, with the last (third) column populated with the {\it refsim} score associated with the sentence. The article excerpt starts with largely true sentences introducing bipolar disorder (sentence \#1) and standard treatments for it (\#2), followed by introducing the main point, that of a dubious probiotics based treatment (\#3) backed up by a spurious study (\#4). Then, as is typical of good hoaxes, it moves to some truthful details about probiotic bacteria (\#5 and \#6) before once again moving back to references to disinformation (\#7 and \#8). It may be noticed from the {\it ReSCo-CC} outputs that disinformation sentences are being identified correctly, and that the {\it refsim} scores (the external scores computed using the refutation document which are unavailable to {\it ReSCo-CC}, and used only in evaluation) also are correlated well with the disinformation in the document. 

Towards analyzing {\it ReSCo-CC} on the ongoing COVID-19 pandemic, we tested on a number of COVID-19 fake news and obtained promising results. For example, in a popular debunked post\footnote{\url{https://www.facebook.com/photo.php?fbid=3156089874411221\&set=a.100975813255991\&type=3}}, {\it ReSCo-CC} correctly identified the key disinformation, that lemon would {\it 'alkalize'} the immune system among the top disinformation sentences. Within our COVID-19 analysis, {\it ReSCo-CC} was found to identify the core disinformation in each case. We hope to do a rigorous COVID-19 analysis as data accumulates within {\it healthnewsreview.org}, our data source for hoax-refutation pairs. 

%post\footnote{\url{https://www.facebook.com/photo.php?fbid=3156089874411221&set=a.100975813255991&type=3&theater}}, {\it ReSCo-CC} correctly identified the key disinformation, that lemon would {\it 'alkalize'} the immune system among the top disinformation sentences. Within our COVID-19 analysis, {\it ReSCo-CC} was found to identify the core disinformation in each case. We hope to do a rigorous COVID-19 analysis as data accumulates. 

\section{Conclusions}

We considered a novel task, that of identifying/scoring key disinformation sentences within long textual articles that are known to be disinformation-laden, motivated by applications to health fake news, and proposed an unsupervised method for the same. Our method makes use of three features, viz., relevance, smoothness and coherence, as well as two cross-sentence assumptions, cohesion and centrality, within a clustering-based construction over sentences in a document. Based on an empirical evaluation over a wide variety of baselines over a task-specific dataset that we curated (to be made public), we illustrate that our method is very effective in identifying core disinformation sentences within disinformation laden health related articles. 

\subsection{Future Work}

In the light of recent and hitherto unseen interest in health disinformation owing to the COVID-19 situation, we are considering further technological avenues of deepening democratic practices in combating disinformation. In particular, we are developing an evidence-based medicine approach towards annotating key disinformation sentences with information obtained through queries over trusted medical databases such as TRIP\footnote{https://www.tripdatabase.com/}, towards allowing users to verify and ascertain the disinformation themselves. We are also considering usage of captions of images that appear embedded within text articles to further improve key disinformation identification; the images and thus captions obviously relate to key elements in the article providing complementary input to improve disinformation assesment. 

\section*{Acknowledgements} Deepak P was partly supported by projects funded by MHRD SPARC (P620) and UKIERI.

%{\footnotesize
\bibliographystyle{ACM-Reference-Format}
\bibliography{acl2020}

\end{document}